\documentclass[sn-mathphys]{sn-jnl}

\usepackage{graphicx}%
\usepackage{multirow}%
\usepackage{amsmath,amssymb,amsfonts}%
\usepackage{amsthm}%
\usepackage{mathrsfs}%
\usepackage[title]{appendix}%
\usepackage{xcolor}%
\usepackage{textcomp}%
\usepackage{manyfoot}%
\usepackage{booktabs}%
\usepackage{algorithm}%
\usepackage{algorithmicx}%
\usepackage{algpseudocode}%
\usepackage{listings}%
\usepackage{natbib}%
\usepackage[capitalise,nameinlink]{cleveref}
\usepackage{subcaption}
\usepackage{todonotes}
\usepackage[utf8]{inputenc}

\newcommand{\lik}{\text{lik}}
\newcommand{\post}{\text{post}}
\newcommand{\probspace}{\mathcal{P}}

\newcommand{\vparam}{\vtheta}
\newcommand{\param}{\theta}

\newcommand{\vnatparam}{\boldsymbol{\lambda}}

\newcommand{\vmeanparam}{\boldsymbol{\mu}}

\newcommand{\dkl}[2]{\mathbb{D}_{\text{KL}}(#1 \, \|\, #2)}

\newcommand\cut[1]{}

\newcommand{\tvlambda}{\widetilde{\boldsymbol{\lambda}}}
\newcommand{\tlambda}{\widetilde{\lambda}}

\newcommand{\squishlist}{
   \begin{list}{$\bullet$}
    { \setlength{\itemsep}{0pt}      \setlength{\parsep}{3pt}
      \setlength{\topsep}{3pt}       \setlength{\partopsep}{0pt}
      \setlength{\leftmargin}{1.5em} \setlength{\labelwidth}{1em}
      \setlength{\labelsep}{0.5em} } }

\newcommand{\squishlisttwo}{
   \begin{list}{$\bullet$}
    { \setlength{\itemsep}{0pt}    \setlength{\parsep}{0pt}
      \setlength{\topsep}{0pt}     \setlength{\partopsep}{0pt}
      \setlength{\leftmargin}{2em} \setlength{\labelwidth}{1.5em}
      \setlength{\labelsep}{0.5em} } }

\newcommand{\squishend}{
    \end{list}  }

{}
{}
{}
{}

\newcommand{\half}{\mbox{$\frac{1}{2}$}}

\newcommand{\real}{\mbox{$\mathbb{R}$}}

\newcommand{\rnd}[1]{\left(#1\right)}
\newcommand{\sqr}[1]{\left[#1\right]}
\newcommand{\crl}[1]{\left\{#1\right\}}
\newcommand{\myang}[1]{\langle#1\rangle}
\newcommand{\myexpect}{\mathbb{E}}

\newcommand{\gauss}{\mbox{${\cal N}$}}

\newcommand{\myvec}[1]{\mbox{$\mathbf{#1}$}}
\newcommand{\myvecsym}[1]{\mbox{$\boldsymbol{#1}$}}

\newcommand{\valpha}{\mbox{$\myvecsym{\alpha}$}}

\newcommand{\vlambda}{\mbox{$\myvecsym{\lambda}$}}

\newcommand{\vtheta}{\mbox{$\myvecsym{\theta}$}}

\newcommand{\va}{\mbox{$\myvec{a}$}}

\newcommand{\vf}{\mbox{$\myvec{f}$}}

\newcommand{\vm}{\mbox{$\myvec{m}$}}

\newcommand{\vx}{\mbox{$\myvec{x}$}}

\newcommand{\vy}{\mbox{$\myvec{y}$}}

\newcommand{\vB}{\mbox{$\myvec{B}$}}

\newcommand{\vI}{\mbox{$\myvec{I}$}}

\newcommand{\vT}{\mbox{$\myvec{T}$}}

\newcommand{\be}{\begin{equation}}
\newcommand{\ee}{\end{equation}}
\newcommand{\bea}{\begin{eqnarray}}
\newcommand{\eea}{\end{eqnarray}}
\newcommand{\beaa}{\begin{eqnarray*}}
\newcommand{\eeaa}{\end{eqnarray*}}

\crefname{section}{Sec.}{Sections}
\crefname{appendix}{App.}{Appendices}
\crefname{algorithm}{Alg.}{Algorithms}
\crefname{equation}{Eq.}{Eqs.}
\crefname{figure}{Fig.}{Figures}
\creflabelformat{equation}{#2\textup{#1}#3} 

\usepackage{thmtools}
\declaretheorem[
mdframed={
  skipabove=6pt,
  skipbelow=6pt,
  hidealllines=true,
  backgroundcolor={lightgray},
  innerleftmargin=8pt,
  innerrightmargin=8pt}
]{ex}

\theoremstyle{thmstyleone}%
\theoremstyle{thmstyletwo}%

\theoremstyle{thmstylethree}%

\begin{document}

\title[]{A Generalization of Amari's Bayesian Duality}


\author[1,2,3]{\fnm{Mohammad Emtiyaz} \sur{Khan}}\email{emtiyaz.khan@riken.jp}

\author[1]{\fnm{Thomas} \sur{M\"ollenhoff}}\email{thomas.moellenhoff@riken.jp}
%
%

\affil[1]{\orgdiv{RIKEN Center for Advanced Intelligence Project}, \orgaddress{\street{1-4-1 Nihonbashi, Chuo-ku}, \city{Tokyo}, \postcode{103-0027}, \state{Tokyo}, \country{Japan}}}
\affil[2]{\orgdiv{Department of Computer Science}, \orgname{Technische Universität Darmstadt}, \orgaddress{\street{Hochschulstraße 10}, \city{Darmstadt}, \postcode{64289}, \state{Hessen}, \country{Germany}}}
\affil[3]{\orgdiv{The Hessian Center for Artificial Intelligence}, \orgaddress{\street{Landwehrstraße 50a}, \city{Darmstadt}, \postcode{64293}, \state{Hessen}, \country{Germany}}}

%


\abstract{Amari's contributions to information geometry and machine learning are well known. Here, we revisit Amari's work on Bayesian duality which has not received as much attention. We connect Amari's Bayesian duality to a convex duality of Bayes' rule. Using this connection, we present a generalization of Amari's Bayesian duality and discuss its relevance for modern artificial intelligence.}

\keywords{Bayes' Rule, Information Geometry, Convex Duality}



\maketitle

\section{Introduction}\label{sec1}
Amari has made many significant contributions to information geometry and machine learning. He was one of the first to use stochastic gradient descent to train neural networks \citep{amari1967theory}. He proposed recurrent neural networks that were earlier versions of what is now called Hopfield networks \citep{amari1979theory}. He also established a connection between \emph{em} and EM algorithms \citep{amari1995information, Am95b} and put
forward a proposal to use natural-gradient methods to train
neural networks \citep{amari1998natural}. All these works (and many others) are now foundational concepts in machine learning and have led to new advances in artificial intelligence.

Here, we revisit a relatively less known work by \citet{amari1996information} on the development of a Bayesian duality theory. This work aimed to explain the `basic but primitive' mechanisms of information processing in the brain and was motivated by the recent works of that time on models such as mixture-of-expert nets, the Helmholtz machine, and the Ying-Yang machine. The work did not receive much attention back then and the focus of the machine-learning community also
partly drifted away from such ideas. Our goal here is to connect Amari's Bayesian duality to other works in the machine-learning literature, so as to generalize its scope and discuss its relevance.

\begin{figure}[t!]
    \centering
    \begin{subfigure}[t]{0.45\linewidth}
        \centering
        \includegraphics[width=\linewidth]{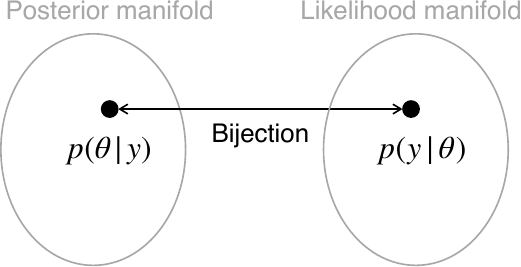}
        \caption{Amari's Bayesian duality}
        \label{fig:fig1a}
    \end{subfigure}
    \hfill
    \begin{subfigure}[t]{0.46\linewidth}
        \centering
        \includegraphics[width=\linewidth]{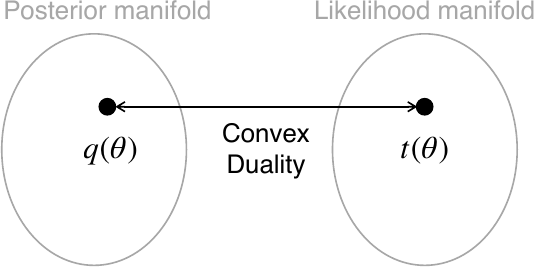}
        \caption{Our generalization}
        \label{fig:fig1b}
    \end{subfigure}

    \caption{\citet{amari1996information} proposed a dual structure connecting the manifolds of posterior and likelihood (Panel \subref{fig:fig1a}). In his case, a likelihood function $p(\vy|\vparam)$ over data vector $\vy$ given parameter vector $\vparam$ is connected to a unique posterior $p(\vparam|\vy)$ and vice versa. His framework relied on a simple posterior form. We present a generalization by using convex duality which not only
       recovers Amari's case but applies much more generally.
    }
    \label{fig:fig1}
\end{figure}

We will start with a description of the basic setup of \citet{amari1996information} where he introduced a new duality structure for the Bayesian framework from the point of view of information geometry. He focused on a specific case where the likelihood and posterior have the same exponential-family (EF) forms and showed that there is a bijection between the two quantities where the roles of their dual coordinates are interchanged (\cref{fig:fig1a}).
The theory however does not apply to the majority of Bayesian cases. For example, in conjugate Bayesian models, the  form of the posterior matches the \emph{prior}, not the likelihood. We will extend the scope of Amari's theory by proposing a more general Bayesian duality through a different mathematical framework that relies on the convex duality of a variational formulation of Bayes' rule (\cref{fig:fig1b}). This will also enable us to connect to a much broader literature in Bayesian inference and convex optimization. We
will conclude the paper by discussing the relevance of Bayesian duality for modern AI.

\section{Amari's Bayesian Duality Theory}

We start with a brief description of Amari's Bayesian duality theory. With a focus on the mechanisms of information processing in the brain, Amari attempted to explain the dynamic interactions between lower and higher systems. We will first discuss
his framework and return to the original motivation again later in the last section.

Amari introduced a new duality structure for a simple case of Bayesian inference by using ideas from information geometry. In his example, we have an observation vector $\vy$ consisting of scalar entries $y_i$ which are connected to the parameter vector $\vparam$ of scalar entries $\param_i$. The two vectors are assumed to be of the same size (although an extension to different lengths was also briefly mentioned). Amari described a duality theory associated with the likelihood function and posterior density, denoted as,
\begin{equation}
   p(\vy\,|\,\vparam) \qquad \text{and} \qquad p(\vparam\,|\,\vy),
\end{equation}
respectively. The notation $p$ is used to denote both densities over $\vy$ and $\vparam$, which is an abuse of notation for simplicity but is standard practice in many Bayesian texts. 

\subsection{The Likelihood Function}
Amari considered a simple case where both the likelihood and posterior take a canonical exponential-family (EF) form. Let us define the likelihood to be 
\begin{equation}
   p(\vy\,|\,\vparam) = h_\lik(\vy) \exp\sqr{ \vparam^\top\vy - A_\lik(\vparam)}.
   \label{eq:lik}
\end{equation}
This is a familiar form: $A_\lik(\vparam)$ is the log-partition function and $h_\lik(\vy)$ is a function of $\vy$, sometimes referred to as the base measure \citep{WainwrightJordan08}. 

The set of such densities is a manifold (denote it by $S_\lik$) with two coordinate systems consisting of natural parameters (denoted by $\vnatparam_\lik$) and expectation parameters (denoted by $\vmeanparam_\lik$), respectively. For the EF in \cref{eq:lik}, we have the following: 
\begin{equation}
   \vnatparam_\lik = \vparam, 
   \qquad
   \vmeanparam_\lik = \myexpect_{p(\text{\vy}|\text{\vparam})}[\vy].
\end{equation}
It is a well-known fact that these two quantities provide two different but equivalent ways to parameterize EFs (for example, see \cite{Ba78}). The coordinates are \emph{dual} of each other due to their relationship through the Legendre transform. Specifically, we have $\vmeanparam_\lik = \nabla A_\lik(\vnatparam_\lik)$ and $\vnatparam_\lik = \nabla A_\lik^*(\vmeanparam_\lik)$ where $A_\lik^*$ is the convex dual of $A_\lik$. In information geometry, this is referred to as the \emph{dually-flat} structure. 

\subsection{The Posterior Distribution}
Now, let us introduce the Bayesian framework where a posterior density is obtained by using the above likelihood along with a prior denoted by $p(\vparam)$. Using Bayes' rule, we can write the posterior density as
\begin{equation}
   p(\vparam\,|\,\vy) = \frac{p(\vy\,|\,\vparam) p(\vparam)}{p(\vy)},
   \label{eq:bayes_rule}
\end{equation}
where the marginal density is denoted by $p(\vy) = \int p(\vy\,|\,\vparam) p(\vparam) d\vparam$.

Amari considered a case where the form of the posterior distribution $p(\vparam\,|\,\vy)$ with respect to $\vparam$ is the same as the form of $p(\vy\,|\,\vparam)$ with respect to $\vy$. This differs from the more popular case of conjugate-Bayes where this is not necessarily the case. There, the form of the posterior matches the form of the \emph{prior}. The likelihood can be expressed in the same form with respect to $\vparam$, but it is rarely the case that is has the same form also with
respect to $\vy$. Amari considered a restrictive case where this is possible by absorbing the prior into the base measure of the likelihood. 
This is shown below where we first substitute the likelihood from \cref{eq:lik}, then rearrange the numerator to write it in a similar form as the likelihood:     
\begin{align}
      p(\vparam\,|\,\vy) &= \frac{ h_\lik(\vy) }{ p(\vy) } \exp\sqr{ \vparam^\top\vy - A_\lik(\vparam)} p(\vparam) \nonumber\\
                     &= \underbrace{ p(\vparam) \exp[-A_\lik(\vparam)] }_{=h_\post(\text{\vparam})} \exp\Big[ \vparam^\top \underbrace{ \vy }_{=\vnatparam_\post} - \underbrace{ \rnd{ \log p(\vy) - \log h_\lik(\vy) } }_{=A_\post(\vy)} \Big]   \nonumber\\
                     &= h_\post(\vparam) \exp\sqr{ \vy^\top\vparam - A_\post(\vy)} . \label{eq:post}
\end{align}
In the last line, we redefined the base measure and log-partition function to write the posterior in the exact same form as the likelihood in \cref{eq:lik}. Note that this rearrangement is only possible if absorbing $p(\vparam)$ still defines a valid EF distribution where $\vy$ is the natural parameter (which is rarely the case as mentioned earlier). However, when it is indeed possible to do so, then, similarly to the likelihood, the posterior also forms a manifold (denote it by $S_\post$) with dual-coordinate systems consisting of the natural and expectation parameters defined below: 
\begin{equation}
   \vnatparam_\post = \vy, 
   \qquad
   \vmeanparam_\post = \myexpect_{p(\text{\vparam}|\text{\vy})}[\vparam].
\end{equation}
Together, the likelihood and posterior give rise to two distinct manifolds.

\subsection{Bayesian Duality}

\citet{amari1996information} showed that the two manifold structures associated with the likelihood and posterior, respectively, are connected to each other through a bijection (\cref{fig:fig1a}) and the roles of the dual coordinates are also interchanged.
This can be more clearly seen in the table below which summarizes the density, natural parameters, sufficient statistics, and expectation parameters, respectively. If all instances of $\vy$ and $\vparam$ are swapped in the top row, we get the bottom row.
\renewcommand{\arraystretch}{1.5}
\begin{table}[!h]
   \begin{tabular}{ccccc}
      \hline
      Name & Density & Nat. param. & Suff. Stat. & Exp. param. \\
      \hline
      Likelihood & $p(\vy|\vparam)= h_\lik(\vy) \exp\sqr{ \vparam^\top\vy - A_\lik(\vparam)}$ & $\vparam$ & $\vy$ & $\myexpect_{p(\text{\vy}|\text{\vparam})}[\vy]$\\
      \hline
      Posterior & $p(\vparam|\vy) = h_\post(\vparam) \exp\sqr{ \vy^\top\vparam - A_\post(\vy)}$  & $\vy$ & $\vparam$ & $\myexpect_{p(\text{\vparam}|\text{\vy})}[\vparam]$\\
      \hline
   \end{tabular}
\end{table}

This connection between the likelihood and posterior is formalized by Amari via the new notion of Bayesian duality. Due to bijection, there is a unique mapping between the two manifolds. Therefore, for a likelihood $p(\vy\,|\,\vparam)$ in $S_\lik$, there exists a unique posterior $p(\vparam\,|\,\vy)$ in $S_\post$, and vice versa. The roles of natural parameters and sufficient statistics are swapped.
Amari referred to this as a new dual structure associated with Bayesian inference, giving rise to a new Bayesian duality theory.

To further illustrate the point, we give a simple example of the dual structure for a case where both likelihood and posterior are isotropic Gaussians.
\begin{ex}
   \label{ex:guassian}
   Consider a Gaussian likelihood function for $\vy$ written in an EF form:
   \begin{equation}
      p(\vy|\vparam) = \gauss(\vy|\vparam, \vI) = \underbrace{ (2\pi)^{-D/2} \exp(-\half \vy^\top\vy) }_{h_\lik(\text{\vy})} \exp\Big[ \vy^\top\vparam - \underbrace{ \half \vparam^\top\vparam}_{A_\lik(\text{\vparam})} \Big],
      \label{eq:iso_gauss_ef}
   \end{equation}
   For this likelihood, the natural parameter is $\vnatparam_\lik = \vparam$ and the expectation parameter is also the same because $\vmeanparam_\lik = \nabla A_\lik(\vparam) = \vparam$. The dual-coordinate pair is simply $ (\vnatparam_\lik, \vmeanparam_\lik) = (\vparam, \vparam)$.

   Let us turn to the posterior density next. We want the posterior to have the same form as the likelihood. A Gaussian prior ensures that the posterior is Gaussian as well, but it does not ensure that the posterior covariance is also identity. However, this happens if we choose a uniform prior. In that case, the marginal density is also uniform, as shown below,
   \begin{equation}
      p(\vy) = \int \gauss(\vy\,|\,\vparam, \vI) d\vparam = \int \gauss(\vparam\,|\,\vy, \vI) d\vparam = 1.
   \end{equation}
   As a result, we can write the posterior as an isotropic Gaussian density:
   \begin{equation}
   p(\vparam\,|\,\vy) = \frac{\gauss(\vy\,|\,\vparam, \vI) p(\vparam)}{ p(\vy)} = \gauss(\vparam\,|\,\vy,\vI).
   \end{equation}
   This also shows that the posterior mean is simply equal to $\vy$. By writing the density in the EF form as in \cref{eq:iso_gauss_ef} we can show that the natural parameter is also equal to $\vy$ and that the posterior takes the same form as the likelihood:
   \begin{equation}
      p(\vparam\,|\,\vy) = \underbrace{ (2\pi)^{-D/2} \exp(-\half \vparam^\top\vparam) }_{h_\post(\text{\vparam})} \exp\Big[ \vparam^\top\vy - \underbrace{ \half \vy^\top\vy}_{A_\post(\text{\vy})} \Big],
      \label{eq:post_iso_gauss_ef}
   \end{equation}
    Therefore, the dual-coordinate pair is $ (\vnatparam_\post, \vmeanparam_\post) = (\vy,\vy)$. As Amari pointed out, the roles of $\vparam$ and $\vy$ are interchanged for manifolds $S_\lik$ and $S_\post$ and there is a trivial bijection between them. 
\end{ex}

\subsection{How to Generalize Amari's Framework?}
\label{sec:how_to}

The main limitation of Amari's Bayesian duality is the restriction that the likelihood and posterior need have the same form. Amari did show an example on the Boltzmann machine but applying it to even simple machine-learning models runs into difficulties. We will now discuss this difficulty for a simple ridge regression problem. We will show that a dual structure can still be found but handling the general cases requires a different mathematical framework. In the next section, we will
present one such framework based on convex duality.

Consider a one-dimensional ridge regression problem with a scalar parameter $\param\in\real$ to model $N$ scalar outputs $y_i\in\real$ given scalar inputs $x_i \in\real$. We denote the vectors of outputs and inputs by $\vy$ and $\vx$ respectively (both of length $N$). We assume a Gaussian likelihood with variance 1 and a standard Gaussian prior:
\begin{equation}
   p(\vy\,|\,\param) = \gauss(\vy\,|\,\vx\param, \vI)
   \qquad \qquad
   p(\param) = \gauss(\param\,|\,0, 1).
   \label{eq:ridge_reg_model}
\end{equation}
With this choice, the posterior is a Gaussian density as well:
\begin{equation}
   p(\param\,|\,\vy) = \gauss(\param\,|\,m_*, \sigma_*^2) \text{ where }  m_* = \sigma_*^2 \vx^\top\vy \text{ and } \sigma_*^2 = 1/(\vx^\top\vx + 1).
   \label{eq:post_oned}
\end{equation}
Unlike the likelihood, the posterior has a variance that is not equal to 1. The likelihood and posterior therefore take different EF forms, and interchanging the roles of the dual coordinates does not make much sense. However, it turns out that we can still find a mapping between the two manifolds. We will first describe the interchanging process and give a formal mathematical framework in the next section.

The key idea is to express the likelihood in terms of the posterior's sufficient statistics. For example, for the posterior in \cref{eq:post_oned}, the sufficient statistic is a 2D vector 
\begin{equation}
   \vT(\param) = (\param, \param^2)^\top,
\end{equation}
and it is therefore possible to write the posterior in an EF form as 
\begin{equation}
   p(\param\,|\,\vy) \propto \exp(\myang{\vnatparam_\post, \vT(\param) }),
\end{equation}
where $\myang{\cdot,\cdot}$ is an inner product and $\vnatparam_\post$ is the natural parameter of the posterior; see \citet[Sec. 5]{khan2023bayesian}. Our idea is to first write the likelihood in the same form, for example, we can find a vector $\tvlambda_\lik$ such that 
\begin{equation}
   p(\vy\,|\,\param) \propto t(\param) = \exp( \myang{\tvlambda_\lik, \vT(\param)}),
\end{equation}
and then write a dual problem as an inverse mapping to find $t(\param)$, giving rise to a dual structure. This is the basic idea of our setup.

We will now show how to write the likelihood in this form and that doing so yields an interchanging of the coordinates, which is similar in spirit to Amari's case.
\begin{ex}
   \label{ex:ridge_exchange}
To show this, we first write the likelihood in the canonical EF form as in the previous example and then redefine the sufficient statistics and natural parameters to write them as an inner product,
\begin{equation}
   \begin{split}
      p(\vy\,|\,\param) &= \underbrace{ (2\pi)^{-\frac{N}{2}} \exp(-\half \vy^\top\vy) }_{h_\lik(\vy)} \exp\Big( \underbrace{\vy}_{\text{\vT}_\lik(\text\vy)} {}^\top \underbrace{ (\vx\param) }_{\vnatparam_\lik} - \underbrace{ \half \vx^\top\vx\param^2 }_{A_\lik(\vnatparam_\lik)}  \Big) \\ 
                    &= h_\lik(\vy) \exp\Bigg[ \left\langle \underbrace{ \begin{bmatrix} \vy\\ 1 \end{bmatrix} }_{\widehat{\text{\vT}}_\lik(\text{\vy})},  \underbrace{ \begin{bmatrix}  \vx\param \\ -\half \vx^\top\vx \param^2 \end{bmatrix}}_{\widehat\vnatparam_\lik} \right\rangle \Bigg] 
\end{split}
\end{equation}
In the second line, we have defined an inner product between $(N+1)$-dimensional vectors where we denote the new sufficient statistics and natural parameter by $\widehat{\vT}_\lik$ and $\widehat{\vlambda}_\lik$, respectively. Next, we write the term inside the exponential as an inner product between 2D vectors which yields a form in terms of $\vT(\param)$ on the right, 
\begin{equation}
   \left\langle \underbrace{ \begin{bmatrix} \vy\\ 1 \end{bmatrix} }_{\widehat{\text{\vT}}_\lik(\text{\vy})},  \underbrace{ \begin{bmatrix}  \vx\param \\ -\half \vx^\top\vx \param^2 \end{bmatrix}}_{\widehat{\vnatparam}_\lik} \right\rangle  
   =  \left\langle \underbrace{ \begin{bmatrix} \vx^\top\vy\\ -\half\vx^\top\vx \end{bmatrix} }_{ \widetilde\vnatparam_{\lik}} ,  \underbrace{ \begin{bmatrix} \param \\ \param^2 \end{bmatrix} }_{ \text{\vT}(\param) } \right\rangle .
   \label{eq:ridge_swap}
\end{equation}
The whole rearrangement can be more compactly written as 
\begin{equation}
   \myang{ \widehat\vT_\lik(\text{\vy}), \widehat\vnatparam_\lik } = \myang{ \tvlambda_\lik , \vT(\param) },
   \label{eq:swap}
\end{equation}
which shows an interchanging of the sufficient statistic and natural parameter.
\end{ex}
\vspace{0.3cm}

The example suggests that, do find a dual mapping, we rewrite Bayes' rule in a form where we replace the likelihood by its mapping $t(\vparam)$ in the space of $\vT(\vparam)$.
\begin{equation}
   p(\vparam\,|\,\vy) \propto t(\vparam) p(\vparam)
\end{equation}
This way the forward map is simply Bayes' rule, while the reverse map is to find $t(\vparam)$ given the posterior $p(\vparam\,|\,\vy)$. We will show next that these operations are written as duals of each other through convex duality. We will use it to generalize Amari's Bayesian duality to general Bayesian cases.

\section{Bayesian Duality via Convex Duality}

We will now derive a dual structure of Bayes' rule by using convex duality. This framework removes the restrictions imposed in Amari's work. We do not claim that this solution is better than using information geometry. Rather, we think that this viewpoint is helpful to connect Amari's work to other works that also use convex duality for Bayesian inference. The key idea is to write Bayes' rule as an optimization problem and then use its dual to find an inverse mapping to search for the
likelihood mapping \smash{$t(\vparam)  = \exp(\myang{\tvlambda, \vT(\vparam)})$} in a dual space.

\subsection{Variational Formulation of Bayes' Rule}

We start by writing a variational form of Bayes' rule, also known as Gibbs' variational principle in information theory. It is well known that Bayes' rule can be written as an optimization problem in the space of all densities (denoted by $\probspace$) over $\vparam$. Consider the following optimization problem over candidate densities $q\in\probspace$:
\begin{equation}
   q_*(\vparam) = \arg \sup_{q\in \probspace} \,\, \myexpect_{q(\text{\vparam})}[\log p(\vy|\vparam)] - \dkl{q(\vparam)}{p(\vparam)},
   \label{eq:var_form}
\end{equation}
where the second term on the right is the Kullback-Leibler divergence (KLD) between the candidate $q(\vparam)$ and the prior $p(\vparam)$. It is well-known that the optimal $q_*$ is the posterior density, that is, $q_*(\vparam) = p(\vparam\,|\,\vy)$. One can also show that the optimal \emph{value} is simply the log-partition function. More precisely, we have 
\begin{equation}
   \log \int p(\vy|\vparam) p(\vparam) d\vparam  = \sup_{q\in \probspace} \,\, \myexpect_{q(\text{\vparam})}[\log p(\vy|\vparam)] - \dkl{q(\vparam)}{p(\vparam)}.
   \label{eq:var_form_val}
\end{equation}
The left hand side is the log-partition function $\log p(\vy)$. The result is easy to verify by simply setting $q = q_*$ on the right-hand side:
\begin{equation}
      \myexpect_{q_*(\text{\vparam})}[\log p(\vy|\vparam)] - \dkl{q_*(\vparam)}{p(\vparam)}  
      = \myexpect_{q_*(\text{\vparam})}\sqr{ \log \frac{p(\vy|\vparam) p(\vparam)}{q_*(\vparam)} } = \log p(\vy).
\end{equation}
This result is attributed to \citet{gibbs1902elementary} and is sometimes referred to as the Gibbs variational principle. The connection to Bayes' rule was formalized by Jaynes, starting from his work on the maximum-entropy principle \citep{jaynes1957}. A general result that goes beyond likelihood functions and applies to generic losses is given in a series of papers by \citet{DoVa75a,DoVa75b,DoVa75c,DoVa75d}.

\subsection{A Dual of the Gibbs Variational Formulation}
Amari's bijection between posterior and likelihood can be derived as a special case of a dual version of the Gibbs variational principle. The goal of the dual problem is to find a likelihood function given a posterior distribution $q(\vparam) \in \probspace$. For this purpose, let us define another space $\mathcal{\probspace}^*$ consisting of valid likelihood functions. For instance, we can choose potential likelihoods $t(\vparam)$ from a function space $\mathcal{F}$ for which the log-partition is finite. Formally, we define
\begin{equation}
   \begin{split}
      \mathcal{\probspace}^* &= 
      \crl{ t \in \mathcal{F} : \log \int t(\vparam) p(\vparam) d\vparam < \infty }.
   \end{split}
\end{equation}
Given such a space, we can write the dual problem as the following:
\begin{equation}
   \dkl{q(\vparam)}{p(\vparam)} = \sup_{t \in \probspace^*} \,\, \myexpect_{q(\text{\vparam})}[\log t(\vparam)] - \log \int t(\vparam) p(\vparam) d\vparam.
   \label{eq:var_form_dual}
\end{equation}
This dual problem recovers the KLD between the posterior $q(\vparam)$ and the prior $p(\vparam)$ by solving an optimization problem over a function space $\probspace^*$. We propose it as a generalization of Amari's Bayesian duality (\cref{fig:fig1b}). We refer to \cref{eq:var_form_dual} as the dual of the Gibbs' variational principle, even though different names are used in the literature.
For instance, in information theory, this is sometimes referred to as the Donsker-Varadhan formula \citep[Thm. 4.6]{polyanskiy2025information}.

For a given $\vy$, if we set $q(\vparam) = q_*(\vparam)$ (the posterior $p(\vparam\,|\,\vy)$ from \cref{eq:var_form}), then the assertion is that an optimal solution $t_*(\vparam)$ recovers the likelihood function $p(\vy\,|\,\vparam)$ up to a constant factor that is absorbed in the normalization. More precisely, when $t_*(\vparam)$ is plugged in Bayes' rule, it will yield the posterior
$p(\vparam\,|\,\vy)$. The equality can be verified by substituting the optimal $t(\vparam) = p(\vy\,|\,\vparam)$ and $q = q_*$,
\begin{equation*}
      \myexpect_{q_*(\text{\vparam})}[\log p(\vy|\vparam)] - \log \int p(\vy|\vparam) p(\vparam) d\vparam  
      = \myexpect_{q_*(\text{\vparam})}\sqr{ \log \frac{p(\vy|\vparam)}{p(\vy)} }
      = \myexpect_{q_*(\text{\vparam})}\sqr{ \log \frac{q_*(\vparam)}{p(\vparam)} }.
\end{equation*}

\subsection{Amari's Bayesian Duality via Convex Duality}

Before giving more details, we connect the above duality to Amari's bijection by applying it to the Gaussian case discussed in \cref{ex:guassian}.
\begin{ex}
   We start by writing the variational form of \cref{eq:var_form}. Because we know that the posterior is an isotropic Gaussian, we can simplify the optimization by restricting it to that space, that is, we set $q(\vparam) = \gauss(\vparam\,|\,\vm, \vI)$. This enables us to write the optimization problem over the mean $\vm$. This is shown below:
   \begin{equation}
      \begin{split}
         \vm_* &= \arg \sup_{\text{\vm}} \,\, \myexpect_{\text{\gauss}(\text{\vparam}\,|\,\text{\vm},\text{\vI})}[\log \gauss(\vy\,|\,\vparam, \vI)] - \dkl{\gauss(\vparam\,|\,\vm,\vI)}{p(\vparam)}\\
               &= \arg \sup_{\text{\vm}} \,\, -\half (\vy-\vm)^\top(\vy-\vm). 
      \end{split}
   \end{equation}
   We see that Bayes' rule in this case reduces to optimization of a quadratic function with maximum $\vm_* = \vy$.

   A similar derivation can be used to write the dual form as a quadratic problem. Mimicking the form of the likelihood, we define $\log t(\vparam)$ to be a quadratic,
   \begin{equation}
      t(\vparam) = \exp\rnd{ \tvlambda^\top\vparam - \half \vparam^\top\vparam },
   \end{equation}
   with $\tvlambda$ as the free parameter. Given a posterior $q_* = \gauss(\vparam\,|\,\vm_*, \vI)$, our goal then is to find the optimal $\tvlambda_*$. The dual problem in \cref{eq:var_form_dual} can then be written as
   \begin{align}
         \tvlambda_* &= \arg \sup_{\tvlambda} \,\, \myexpect_{\text{\gauss}(\text{\vparam}|\text{\vm}_*,\text{\vI})}[\log t(\vparam)] - \log \int t(\vparam) p(\vparam) d\vparam \notag \\
               &= \arg \sup_{\tvlambda} \,\, \myexpect_{\text{\gauss}(\text{\vparam}|\text{\vm}_*,\text{\vI})} \left[ \tvlambda^\top\vparam - \half \vparam^\top\vparam \right] - \log \int \exp\rnd{ \tvlambda^\top\vparam - \half \vparam^\top\vparam } p(\vparam) d\vparam \notag\\
               &= \arg \sup_{\tvlambda} \,\, \tvlambda^\top\vm_* - \half \vm_*^\top\vm_* - \frac{D}{2} \log(2\pi) - \half \tvlambda^\top\tvlambda \notag \\
               &= \arg \sup_{\tvlambda} \,\, \tvlambda^\top\vm_* - \half \tvlambda^\top\tvlambda.\notag  
   \end{align}
   This is a quadratic function whose solution is $\tvlambda_* = \vm_* = \vy$. Therefore, we have $t_*(\vparam) \propto \gauss(\vy\,|\,\vparam, \vI)$. We note that the function $t(\vparam)$ need not be a probability density, but its role is identical. If we replace $p(\vy\,|\,\vparam)$ in \cref{eq:var_form} by $t_*(\vparam)$, then we recover the posterior $q_*(\vparam)$ which is also equal to the posterior $p(\vparam\,|\,\vy)$.
\end{ex}
\vspace{0.3cm}
This example illustrates that the primal-dual formulation of the Gibbs' variational principle can realize the bijection proposed by Amari when we use the specific form of the likelihoods and posterior. This dual version however applies much more generally. In fact, it applies even when likelihoods are replaced by generic loss functions. We will now show an application to the ridge regression case. 

\subsection{Bayesian Duality of Ridge Regression}

We return to the example of ridge regression shown in \cref{eq:ridge_reg_model}. Amari's framework is difficult to apply to this case and we will show that by using the dual problem over $t(\param)$, we can solve this issue. We showed in \cref{eq:ridge_swap} of \cref{ex:ridge_exchange} that the optimal likelihood mapping is a 2D vector: \smash{$\tvlambda_\lik = (\vx^\top\vy, -\half \vx^\top\vx)^\top$}. We will now show that this solution can be recovered by solving the dual problem.
\begin{ex}
   We restrict $\log t(\param)$ to be a quadratic function with two scalar parameters $\tlambda_1$ and $\tlambda_2$, as shown below:
   \begin{equation}
      t(\param) = \exp\rnd{ \tlambda_1\param - \half \tlambda_2\param^2 }.
   \end{equation}
   Given a posterior $q_* = \gauss(\param\,|\,m_*, \sigma_*^2)$, our goal is to find the optimal $\tlambda_{1,*}$ and $\tlambda_{2,*}$. The dual problem in \cref{eq:var_form_dual} can be written as
   \begin{equation}
      \begin{split}
         \tvlambda_* &= \arg \sup_{\tvlambda} \,\, \myexpect_{\text{\gauss}(\param|m_*,\sigma_*^2)}[\log t(\param)] - \log \int t(\param) p(\param) d\param\\
               &= \arg \sup_{\tvlambda} \,\, \tlambda_1 m_* - \half \tlambda_2 (m_*^2 + \sigma_*^2) - \log \int \exp\rnd{ \tlambda_1\param - \half \tlambda_2\param^2  } p(\param) d\param.
      \end{split}
      \label{eq:bayesdualobj_ridge}
   \end{equation}
   The integral is available in closed form:
   \begin{equation}
      \log \int \exp\rnd{ \tlambda_1\param - \half \tlambda_2\param^2  } p(\param) d\param = - \half \log (\tlambda_2 + 1) + \half \frac{\tlambda_1^2}{\tlambda_2 + 1}.
      \label{eq:integral}
   \end{equation}
   Substituting this in the optimization problem and taking derivatives, we get
   \begin{equation}
      \tlambda_{1,*} = m_*/\sigma_*^2 = \vx^\top\vy, \qquad\qquad \tlambda_{2,*} = 1/\sigma^2_* -1 = \vx^\top\vx,
      \label{eq:opt_ex4}
   \end{equation}
   which is the desired solution.
   A detailed derivation is in \cref{app:deriv}.
   \label{ex:ridgedual}
\end{ex}
\vspace{0.3cm}

The dual problem yields the optimal mapping for the likelihood in the $\vT(\param)$ space. Using the interchanging operation shown in \cref{eq:swap} it may be possible to then write an appropriate likelihood. Unlike Amari's example, this operation will not be a bijection in general because there could be several ways to aggregate the information. 

It is also possible to write the dual problem directly in the data space. 
For instance, we can rewrite the dual problem in the space of the data vector $\vy$ which is of length $N$. This is the space of linear predictions denoted by
\begin{equation}
   \vf = \vx \param.
   \label{eq:fspace}
\end{equation}
This formulation can recover the likelihood exactly, as shown next.

\begin{ex}
   \label{ex:functionspace}
   The dual problem can be written in the $\vf$-space by using a pushforward of the distribution. Specifically, we can write the prior and posterior in the $\vf$-space using a change of variable.
   \begin{equation}
      p(\vf) = \gauss(\vf\,|\,0, \vx\vx^\top),
      \qquad\qquad
      q_*(\vf) = \gauss(\vf\,|\,\vx m_*, \sigma_*^2\vx\vx^\top).
   \end{equation}
   The prior is degenerate but that does not pose a problem as long as the integrals with respect to it are finite.
   The likelihood candidate can also be mapped in that space by redefining the free parameters to be of appropriate sizes:
   \begin{equation}
      t(\vf) = \exp\rnd{ \va^\top\vf- \half \vf^\top\vB\vf },
      \label{eq:site}
   \end{equation}
   where $\va$ is an $N$-length real vector while $\vB$ is an $N\times N$ real matrix.
   With these, the dual problem can be written in the $\vf$-space, as an optimization over $\va$ and $\vB$, 
   \begin{align}
         (\va_*,&\vB_*) = \arg \sup_{\text{\va},\text{\vB}} \,\, \myexpect_{q_*(\text{\vf})}[\log t(\vf)] - \log \int t(\vf) p(\vf) d\vf \label{eq:bdfull} \\
                    &= \arg \sup_{\text{\va},\text{\vB}} \,\, \va^\top\vx m_* - \half \vx^\top\vB\vx (m_*^2 + \sigma_*^2) - \log \int \exp\rnd{ \va^\top\vf- \half \vf^\top\vB\vf } p(\vf) d\vf. \nonumber         
   \end{align}
   The integral in the last term can be obtained by changing back the variable to $\param$ and then substituting in \cref{eq:integral} the following: $\tlambda_1 = \va^\top\vx$ and $\tlambda_2 = \vx^\top\vB\vx$. Using this, we can write the optimization problem as follows:
   \begin{align}
      \va^\top\vx m_* - \half \vx^\top\vB\vx (m_*^2 + \sigma_*^2) + \half \log (\vx^\top\vB\vx + 1) - \half \frac{(\va^\top\vx)^2}{\vx^\top\vB\vx + 1}.
   \end{align}
   Taking the derivatives, we get the optimality condition that is identical to \cref{eq:opt_ex4},
   \begin{equation}
      \vx^\top\va_* = m_*/\sigma_*^2 = \vx^\top\vy, \qquad\qquad \vx^\top\vB_*\vx = 1/\sigma^2_* -1 = \vx^\top\vx.
   \end{equation}
   These are satisfied by $\va_* = \vy$ and $\vB_* = \vI$, so a solution recovers the likelihood, 
   \begin{equation}
      t_*(\vf) = \exp\rnd{ \vy^\top\vx\param- \half \vx^\top\vx\param^2 }
      \,\propto\, \gauss(\vy\,|\,\vx\param, \vI)
      = p(\vy\,|\,\param).
   \end{equation}
\end{ex}
\vspace{0.3cm}

This dual formulation yields the likelihood in $p(\vy\,|\,\param)$ space, but since the dual solution is not unique, there is no bijection between posterior and likelihood. The formulation automatically performs the interchanging operation shown in \cref{eq:swap} similarly to Amari's case. 

\section{Relevance for AI and Connections to Other Fields}

So far, we have shown that Amari's Bayesian duality can be generalized through convex duality where the dual problem can be seen as an inverse (and sometimes bijective) mapping from the posterior manifold to the likelihood manifold. But, why is this relevant for machine learning and AI?

\begin{figure}[t!]
    \centering
    \begin{subfigure}[t]{0.48\linewidth}
        \centering
        \includegraphics[width=\linewidth]{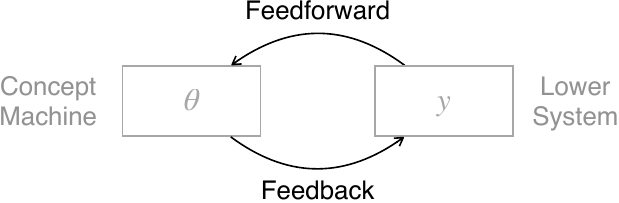}
        \caption{}
        \label{fig:fig2a}
    \end{subfigure}
    \hfill
    \begin{subfigure}[t]{0.45\linewidth}
        \centering
        \includegraphics[width=\linewidth]{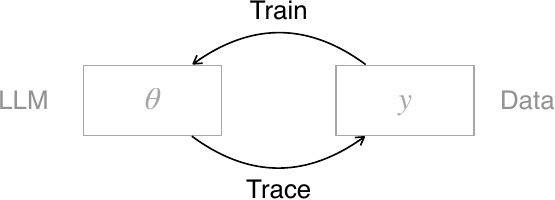}
        \caption{}
        \label{fig:fig2b}
    \end{subfigure}

    \caption{Amari's motivation was to explain the dynamic interaction between a higher-order system (concept machine) and a lower order system, through an iterative feedforward and feedback mechanism. In modern times, Bayesian duality could be useful to study the relationship between the model and data. For example, the dual problem can be used to \emph{trace} the knowledge of an LLM.
    }
    \label{fig:fig2}
\end{figure}

Amari's original proposal was focused on information processing in the brain. At the time, he aimed to develop a new theory of ``dynamic interaction between a lower and a higher neural systems'' (\cref{fig:fig2a}). The lower order system, modeled by $\vy$, is connected to a sensory input while the higher order system is modeled by
$\vparam$ which is connected to a `concept' machine. He imagined Bayesian duality as a stochastic version of the autoassociative-memory model where the concepts are encoded as binary vectors in the higher system. When a new sensory input is shown, the concept machine uses a `feedforward' process to estimate a concept that best explains it. This is the inference part to get $p(\vparam|\vy)$. Then, the higher neural system stimulates the lower one by proposing a new $\vy$ through the feedback
process. This is realized by $p(\vy|\vparam)$. This dynamic interaction was the motivation for Amari
to develop Bayesian duality, which he proposed to implement through an \emph{e-m} procedure.

For us too, the interaction between $\vy$ and $\vparam$ is the most attractive part of Bayesian duality. In our research, we are interested in Bayesian duality as a tool to study the relationship between the model and data. For example, given a Large Language Model (LLM) trained on a large data, we would like to understand the model's knowledge. This question can be framed as the inverse mapping: we want to find a compact set of data examples (or prompts) that
can summarize the main crux of the LLM's knowledge. The dual problem can be used to address such problems: given $p(\vparam|\vy)$, the mapping $t(\vparam)$ in the likelihood space can be used to `trace' a compact summary of the model's knowledge (\cref{fig:fig2b}). Amari himself was interested in understanding the underlying concepts encoded in $\vparam$, which he modeled through a subspace in the manifold spanned by a smaller set of coordinates. 

Our work here connects Amari's ideas to other works that rely on convex duality to find compact representations. One of the earliest works on this is the representer theorem proposed by \citet{KiWa70} in a Bayesian context. Later, this led to the representer theorem used in kernel methods \citep{ScHe01} and the support vector machine (SVM) where compact representations in the data space are found by solving an efficient convex-optimization problem. A representer theorem for Gaussian
process models was proposed in \citet[Lemma 1]{csato2002sparse}. General versions of such dualities have also been
studied extensively, for example, by \citet{altun2006unifying} for function spaces and by \citet{dudik2007maximum} to generalize the maximum entropy principle. An SVM-style generalization of Bayesian inference is proposed by \citet{zhu2014bayesian}. We also note that an informal mention of Bayesian duality is made in \citet{robert2007bayesian} for Bayes' rule as an inversion but Amari's usage is more precise. More recently, such dualities are used to obtain primal-dual adversarial interpretations of Bayes \citep{husain2022adversarial, samasbayes}.

In fact, Bayesian duality can recover some non-Bayesian applications of convex duality in machine learning. For instance,
the dual problem for ridge regression of \cref{eq:ridge_reg_model} is a classical reformulation of minimization over $\param$ as a maximization over an $N$-length real-valued dual vector $\valpha$~\citep{rockafellar1967duality}. This is shown below,
\begin{equation}
   \min_{\param} ~ -\log p(\vy|\param) - \log p(\param) \quad \Longleftrightarrow \quad \max_{\text{$\valpha$}} ~ \vy^\top\valpha - \half \valpha^\top (\vI + \vx\vx^\top) \valpha.
   \label{eq:ridgedual}
\end{equation}
We can show that, for this case, the optimal dual vector $\valpha_*$ is related to $t_*(\vf_*)$ in \cref{eq:site} as follows,
\begin{equation}
   \valpha_* = \vy - \vf_* = \nabla \log t_*(\vf_*),
   \label{eq:dual_as_res}
\end{equation}
where $\vf_* = \vx\param_*$ is the optimal prediction. A proof is given in \cref{eq:connection_to_convex_dual}. In general, such equivalences between the non-Bayesian and Bayesian dualities are to be expected due to their common origins in convex duality. We also expect Bayesian duality with more expressive posterior forms to contain as special cases other dualities based on less flexible posteriors and also those that arise in non-Bayesian scenarios.

Recent works have tried to build such versions of Bayesian dualities by using variational approximations. Such methods use dual structures associated with simple posterior approximations (such as Gaussian approximations to the posterior). This makes it possible to derive practical algorithms that can exploit the benefits of Bayesian duality. One such early work is by \citet{KhanAFS13} which proposed a dual formulation for Gaussian variational inference. More recently, \citet{adam2021dual} proposed the use of such dual variables for
approximate inference in Gaussian processes. \citet{khan2025knowledge} uses dual representations to unify knowledge-adaptation methods, while \citet{mollenhoff2026federated} used a dual structure to generalize ADMM for federated learning. The work started by Amari on Bayesian duality is now finally having an impact on modern AI and we hope that it will continue to benefit the community.




\backmatter

%
%
%
%
\bmhead{Acknowledgments}
This work is supported by JST CREST Grant Number JPMJCR2112. We thank many current and past members of the ABI team and the Bayes-duality project for discussions over the years which played an important role in shaping the ideas presented in this paper.


\appendix 

\section{Appendix}
\subsection{Derivation of the Dual Problem for Ridge Regression} 
\label{app:deriv}
To evaluate the integral in \Cref{eq:integral}, we use a change of variables and the fact that $\int \exp(-z^2/2) \mathrm{d}z = \sqrt{2\pi}$. First, we bring the integral into a form amenable to the change of variables.
\begin{align}
   \int &\exp\rnd{ \tlambda_1\param - \half \tlambda_2\param^2  } p(\param) d\param = \frac{1}{\sqrt{2\pi}} \int \exp\rnd{ \tlambda_1\param - \half (\tlambda_2 + 1)\param^2  } d\param \\
   &= \frac{1}{\sqrt{2\pi}} \exp \rnd{\frac{\tlambda_1^2}{2(\tlambda_2 + 1)} } \int \exp\rnd{ -\half (\tlambda_2 + 1) \rnd{\theta - \frac{\tlambda_1}{\tlambda_2 + 1}}^2  } \mathrm{d} \theta.  \label{eq:last}
\end{align}
Then, setting $z = \sqrt{\tlambda_2 + 1}\rnd{\theta - \frac{\tlambda_1}{\tlambda_2 + 1}}$, $\mathrm{d}\theta = \frac{\mathrm{d}z}{\sqrt{\tlambda_2+1}}$ gives, 
\begin{align}
   \int \exp\rnd{ -\half (\tlambda_2 + 1) \rnd{\theta - \frac{\tlambda_1}{\tlambda_2 + 1}}^2  } \mathrm{d} \theta = \frac{1}{\sqrt{\tlambda_2+1}} \underbrace{\int \exp \rnd{-\frac{z^2}{2}} \mathrm{d}z}_{=\sqrt{2\pi}}.
\end{align}
Substituting that back into \Cref{eq:last} and taking the log gives, 
\begin{align}
   \log \frac{1}{\sqrt{\tlambda_2 + 1}}\exp \rnd{\frac{\tlambda_1^2}{2(\tlambda_2 + 1)} } = -\half \log (\tlambda_2 + 1) + \half \frac{\tlambda_1^2}{\tlambda_2 + 1}.
\end{align}
This is the result shown in the main paper. We now differentiate the full objective in \Cref{eq:bayesdualobj_ridge} with respect to $\tlambda_1$ and $\tlambda_2$ and set the resulting derivatives to zero.
\begin{align}
   m_* - \frac{\tlambda_1}{\tlambda_2 + 1} = 0, \qquad 
   -m_*^2 - \sigma_*^2 + \frac{1}{\tlambda_2 + 1} + \frac{\tlambda_1^2}{(1 + \tlambda_2)^2} = 0.
\end{align}
Solving the first equation yields $\tlambda_1 = m_* (\tlambda_2 + 1)$. Inserting that into the equation on the right, we arrive at 
\begin{align}
   \sigma_*^2 = \frac{1}{\tlambda_2 + 1} \quad \Leftrightarrow \quad \tlambda_2 = \frac{1}{\sigma_*^2} - 1. 
\end{align}
Plugging that back into the expression for $\tlambda_1$, we get $\tlambda_1 = m_* / \sigma_*^2$. Since $m_* = \sigma_*^2 \vx^\top \vy$ as per \Cref{eq:post_oned}, we have $\tlambda_1 = \vx^\top \vy$. Since $\sigma_*^2 = 1 / (\vx^\top \vx + 1)$, we get $\tlambda_2 = \vx^\top \vx$ as claimed.

%

\subsection{Connection to Convex Dual of Ridge Regression} 
\label{eq:connection_to_convex_dual}

The primal solution is just the ridge solution which has the following form:
\begin{equation}
   \param_* = (\vx^\top \vx + 1)^{-1} \vx^\top\vy
\end{equation}
The dual solution is obtained from the stationarity condition of the dual problem:
\begin{equation}
   \vy - (\vI + \vx\vx^\top)\valpha_* = 0
   \qquad\implies \qquad
   \valpha_* = (\vx\vx^\top + \vI)^{-1} \vy.
\end{equation}
From here, we can represent $\param_*$ in terms of $\valpha_*$ by using the matrix inversion lemma,
\begin{equation}
   \param_* = (\vx^\top\vx + 1)^{-1} \vx^\top\vy =  \vx^\top(\vx\vx^\top + \vI)^{-1}\vy = \vx^\top\valpha_*
\end{equation}
Using this, we can prove \cref{eq:dual_as_res} from the stationarity condition of the dual problem:
\begin{equation}
   \vy - (\vI + \vx\vx^\top)\valpha_* = 0 
   \quad\implies\quad \vy - \valpha_* - \vx\param_* = 0
   \quad\implies\quad \valpha_* = \vy- \vf_*.
\end{equation}
This is equivalent to 
\begin{equation}
   \nabla \log t_*(\vf_*) = \nabla (\va_*^\top\vf_* - \half \vf_*^\top\vB_* \vf_*) = \va_* - \vB_*\vf_* = \vy-\vf_*. 
\end{equation}

\vspace{.5cm}

\noindent
\textbf{Conflict of Interest: } One of the authors has co-authored several papers with Dr. Frank Nielsen. There are no other obvious conflicts of interest to report at this moment.

\bibliography{refs}

\end{document}